# Chapter 1
# Machine Learning for Soccer Match Result Prediction

Rory Bunker, Calvin Yeung, and Keisuke Fujii

**Abstract**   Machine learning has become a common approach to predicting the outcomes of soccer matches, and the body of literature in this domain has grown substantially in the past decade and a half. This chapter discusses available datasets, the types of models and features, and ways of evaluating model performance in this application domain. The aim of this chapter is to give a broad overview of the current state and potential future developments in machine learning for soccer match results prediction, as a resource for those interested in conducting future studies in the area. Our main findings are that while gradient-boosted tree models such as CatBoost, applied to soccer-specific ratings such as pi-ratings, are currently the best-performing models on datasets containing only goals as the match features, there needs to be a more thorough comparison of the performance of deep learning models and Random Forest on a range of datasets with different types of features. Furthermore, new rating systems using both player- and team-level information and incorporating additional information from, e.g., spatiotemporal tracking and event data, could be investigated further. Finally, the interpretability of match result prediction models needs to be enhanced for them to be more useful for team management.

Rory Bunker
Graduate School of Informatics, Nagoya University, Furo-cho, Chikusa ward, Nagoya, 464-8601, Japan, e-mail: `rory.bunker@g.sp.m.is.nagoya-u.ac.jp`

Calvin Yeung
Graduate School of Informatics, Nagoya University, Furo-cho, Chikusa ward, Nagoya, 464-8601, Japan, e-mail: `yeung.chikwong@g.sp.m.is.nagoya-u.ac.jp`

Keisuke Fujii
Graduate School of Informatics, Nagoya University, Furo-cho, Chikusa ward, Nagoya, 464-8601, Japan, e-mail: `fujii@i.nagoya-u.ac.jp`

## 1.1 Introduction

Predicting the results of professional soccer[1] matches is a challenging problem due to draws being a common outcome in the sport, as well as its low-scoring nature and often highly competitive leagues. Nonetheless, given the global popularity of the sport, both in terms of spectatorship and player numbers, it is a topic that is of interest to many groups including fans, bookmakers and bettors, as well as coaches, players, and performance analysts. While bettors and bookmakers require models that are highly accurate, coaches/management and sports performance analysts also require models that are interpretable, so that the most relevant match features (performance indicators [63]) for winning can be identified and improved upon in future matches. Once a predicted result for a specific match is obtained, an additional problem is to decide whether to actually bet on the match. While this is an important question [38], it is not the focus of the current chapter.

A large number of papers have been published that are related to machine learning (ML) for soccer match result prediction, particularly over the past decade [18]. Traditionally, however, statistical models were used to forecast soccer match results. Stefani [105] used least-squares regression to calculate team ratings, which could be updated on a weekly basis, to predict match results based on the difference in ratings between teams. Some early papers fit distributions to the number of goals scored by each team in a match. Maher [81] used an independent Poisson distribution to obtain the attacking and defensive ratings of teams. Dixon & Coles [37] modified this model to handle incomplete data and data from different divisions, and to allow for temporal variations in the performance of teams. Goals distributions have also been fit using, e.g., the dependent Poisson, negative binomial, and extreme value distributions [51, 9, 10, 91]. Indeed, statistical models such as the Bivariate Poisson still provide strong performance (e.g., [78] who used the engsoccerdata package [34]).

Constantinou [28] categorized soccer result prediction models into three groups, suggesting that the first and third types of studies tend to be published in statistical journals, while the second group tend to be published in journals focused on computer science and artificial intelligence.

1. **Statistical models**: Including modeling goals scored, e.g., using the Bivariate, independent and dependent Poisson, negative binomial, and generalized extreme value distributions [81, 37, 51, 9, 10, 91], as well as ordered logistic regression models [4, 100].
2. **Machine learning and probabilistic graphical models**: Including fuzzy or genetic algorithms [112, 87, 54, 95] and Bayesian methods (e.g., the studies of Joseph, Fenton & Neil [68] and Constantinou, Fenton, & Neil [31]).
3. **Rating systems**: Including Elo ratings [41, 65], pi-ratings [30], and more recently, Berrar ratings [8] and Generalized Attacking Performance (GAP) ratings [118]. PageRank ratings [15, 59] also fit under this category.[2]

---

[1] Also known as "Association football" or simply "football".

[2] How these different ratings are calculated is described in subsection 1.4.1.

Nonetheless, this three-group model taxonomy is not entirely satisfactory. In recent years, researchers have incorporated models, techniques, or features from more than one of the above three groups to create hybrid approaches. For instance, the top-performing participants in the 2017 Soccer Prediction Challenge [7] applied machine learning models [8, 59, 28] to soccer-specific ratings. Statistical and machine learning models have also been combined to form hybrid approaches [40, 74, 53], with [53], for example, proposing a hybrid approach combining a Random Forest model with Poisson-based rankings. In addition, some models, e.g., Logistic Regression, fit under both the umbrella of statistics as well as that of machine learning. On the other hand, Bradley-Terry models [11], which also proved successful during the 2017 Soccer Prediction Challenge [116] and for ranking soccer teams based on current strength [78], could be considered to fit under both the statistical models and rating systems groups. This is because, as will be explored later in this chapter in subsection 1.4.1.1, Bradley-Terry forms the theoretical foundation for Elo ratings [33].

By covering available datasets — including an in-depth discussion of the Open International Database and the 2017 Soccer Prediction Challenge — as well as current and potential future models and features, as well as evaluation methods, this chapter aims to provide a broad overview of machine learning for soccer match result prediction and will hopefully act as a resource for those interested in carrying out future research in the domain.

The remainder of this chapter is organized as follows. In the next section, available datasets are outlined, as are the Open International Soccer Database and 2017 Soccer Prediction Challenge's in-competition and post-competition approaches and results. Subsequently, commonly used candidate models including conventional machine learning models, ensemble methods, and deep learning models are considered, as are model objectives and interpretability. Then, current and potential future model features are considered, e.g., match features, player and team statistics, and ratings (both general and soccer-specific). Feature selection methods are also briefly discussed. Model evaluation considerations — including baselines, target variable definition, as well as scoring rules and their properties, along with temporal splitting methods — are then described. Finally, conclusions and potential avenues for future research are discussed.

## 1.2 Data

In subsection 1.2.1, a non-exhaustive list of the datasets that are available for engineering features and building models is described. The dataset used in the 2017 Soccer Prediction Challenge [7], the Open International Soccer Database [39], will then be discussed, which — despite not containing match features derived from in-play events during games (apart from goals scored) — now acts as somewhat of a benchmark dataset. The top-performing in-competition and post-competition studies that have used the Open International Soccer Database will be discussed in subsection 1.2.2.

### 1.2.1 Available Datasets

An increasing amount of publicly available data from soccer matches is becoming publicly available online. Organizations such as StatsBomb (Bath, United Kingdom) make some event log data publicly available, although the data is generally held by professional teams who pay vendors such as StatsBomb or Stats Perform (Chicago, IL, USA)/Opta (London, United Kingdom) for access. A non-exhaustive list of datasets that can be used for engineering features and building soccer match result prediction ML models is presented in Table 1.1.

One challenge with many of these data sources is that they often cover different leagues and/or different seasons, and may be missing some types of features. The Open International Soccer Database, for instance, does not contain match features apart from goals scored, nor does it contain betting odds. It may be necessary for researchers to source data from different datasets and merge them. The European Soccer Database on Kaggle has already performed such merging for a variety of feature types for a subset of their entire dataset (10,000 of the 25,000 matches). Spatiotemporal data is often only available to professional teams themselves, although sometimes small amounts of it are made publicly available.

### 1.2.2 2017 Soccer Prediction Challenge

For the 2017 Open International Soccer Prediction Challenge, the Open International Soccer Database [39], an open-source database containing over 216,000 matches from 52 soccer leagues and 35 countries, was made available to participants and remains publicly available. As mentioned, this dataset now acts somewhat as a benchmark dataset in this domain despite not containing match features other than goals scored. The omission was likely a deliberate design choice by the challenge organizers, who likely integrated only data that are readily available for most soccer leagues worldwide — including lower-division leagues — so as to maximize the size of the dataset. The papers related to the top-performing models from the 2017 Soccer Prediction Challenge were published in a special issue of the Machine Learning (Springer) Journal. Challenge participants trained models using the 216,000+ match training dataset to predict 206 future match results. The Ranked Probability Score (RPS) [42, 29], which measures how good forecasts are compared to observed outcomes when the forecasts are expressed as probability distributions, was used as the evaluation metric in the competition. Specifically, the participants' models were evaluated based on the RPS averaged across the 206-match challenge test set.

**Table 1.1** Datasets that can potentially be used for ML for soccer match result prediction. The attributes for each dataset can be confirmed by navigating to the corresponding URL.

| Dataset Name | Description | URL(s) |
|---|---|---|
| European Soccer Database | Contains data on 25,000 matches and 10,000 players in 11 European leagues from 2008 – 2016. Player and team attributes are from the EA Sports FIFA video game. The database also contains team lineups and formations (x,y co-ordinates), as well as betting odds from up to 10 bookmakers. For over 10,000 matches, the database also contains events, e.g., goal types, possession, corners, crosses, fouls, and cards. | kaggle.com/datasets/hugomathien/soccer |
| Betting websites | Betting sites provide betting odds and some-times additional features (but sometimes only short-time periods are available for free). | football-data.co.uk, betfair-times.github.io, and oddsportal.com |
| engsoccerdata | An R package [34] that provides English and other European league data, along with US MPL and South African league data. | github.com/jalapic/engsoccerdata |
| Wyscout event data | Pappalardo et al. [90] and Wyscout provided spatiotemporal event data containing 1,941 matches from the top 5 European Soccer leagues, EURO 2016 and the 2018 World Cup. | doi.org/10.6084/m9.figshare.c.4415000.v5 |
| Statsbomb open data | Primarily event log data (with lineup and match metadata). | github.com/statsbomb/open-data |
| Open International Soccer Database | Contains the match results (season, league, date, home team, away team(Sea Lge Date HT AT HS AS GD WDL) of 216,743 games played between 19/03/2000 and 21/03/2017 from 52 leagues in 35 countries. This dataset was used in the 2017 Soccer Prediction Challenge. | osf.io/kqcye/ |
| soccerdata | A scraper collection implemented in Python that scrapes soccer data from various websites including Club Elo, ESPN, FBref, FiveThir-tyEight, Football-Data.co.uk, SoFIFA, and WhoScored. | github.com/probberechts/soccerdata |
| World Football Elo Ratings | Contains the current Elo ratings of national soccer teams. | eloratings.net |
| Football Database | Contains the current Elo ratings of club teams from various leagues around the world. | footballdatabase.com |
| understat.com | Contains expected goals (xG) (however, the way in which these are computed is relatively opaque). | understat.com |
| football-data.co.uk | Contains, for various leagues, betting odds from multiple providers as well as match statis-tics for some seasons. | football-data.co.uk |
| FIFA Index | Contains player and team ratings from the EA Sports FIFA video game | fifaindex.com |

### 1.2.2.1 2017 Soccer Prediction Challenge: Top Performers

We now cover three of the four papers that achieved the best performance in the competition.[3]

---

[3] The fourth top-performing paper [116] is not covered in this chapter because the methods used were more statistical in nature: e.g., a Bradley–Terry model, Poisson log-linear hierarchical model, and integrated nested Laplace approximation.

Hubáček, Sourek & Zelezny [59] used relational- and feature-based methods to build models using the Open International Soccer Database. Pi-ratings and PageRank ratings were computed for each of the teams in each match. Both the regression and classification forms of XGBoost [26] were employed as the feature-based method, while boosted relational dependency networks (RDN-Boost) [85] were used as the relational method. The classification with XGBoost on the pi-ratings feature set performed best on both the validation set and the unseen challenge test set, achieving 0.5243 accuracy and an average RPS of 0.2063. Adding more model features, weighting aggregated data according to recency, and including expert guidance by using, e.g., active learning, were mentioned as possible avenues for further work.

Constantinou [28] created a model that combined dynamic ratings with a Hybrid Bayesian Network. The rating system used was a modified version of pi-ratings, which the author had proposed in previous work [30]. The computation of pi-ratings emphasized the match result (win, draw, or loss) to a greater degree than the goal margin, with the aim of dampening the influence of large goal differences. In contrast to the original pi-ratings system, this version also incorporated a team form factor that aims to identify continued over- or under-performance. Four rating features (two for the home team and two for the away team) were used as the inputs to the Hybrid Bayesian Network. Notably, the proposed Hybrid Bayesian Network with modified pi-ratings was able to effectively predict a match between two teams, even when the prediction was based on historical match data that involved neither of the two teams. On the challenge test set, accuracy of 0.5146 and an average RPS of 0.2083 was obtained. Incorporating other key features or expert knowledge, e.g., player transfers, key player availability, international competition participation, management, injuries, attack/defence ratings, and team motivation/psychology, were mentioned as potential opportunities for future improvement.

Berrar et al. [8] applied two ML models to two different feature sets: recency features and rating features. Recency features were computed by averaging feature values over the previous nine matches and were based on four feature groups: attacking strength, defensive strength, home advantage, and opposition strength. A new soccer-specific rating system was proposed, subsequently referred to as "Berrar ratings."[4] XGBoost [26] and k-Nearest-Neighbors (k-NN) were applied to each of the two feature sets, with both models performing better on the rating features compared to the recency feature set. XGBoost applied to the rating features provided the best performance with accuracy of 0.5194 and an average RPS of 0.2054; however, this result was obtained after the challenge had concluded.[5] The authors noted that soccer's low-scoring nature and narrow margins of victory make it challenging to predict based only on goals, while also emphasizing the importance of effective feature engineering and incorporation of domain knowledge.[6] The authors also suggested including match features (e.g., yellow/red cards, fouls, possession, passing and running rates), player-level characteristics (e.g., salaries, ages, and physical

---

[4] Please refer to subsection 1.4.1.3 for a description of how Berrar ratings are calculated.

[5] k-NN applied to the rating features achieved 0.5049 accuracy and an average RPS of 0.2149 on the challenge test set, which was the best result achieved during the competition itself.

[6] Incorporation of domain knowledge is discussed in the Appendix to this chapter.

**Table 1.2** Ranked probability scores (RPS) and accuracy results from the 2017 Soccer Prediction Challenge and subsequent studies that have used the Open International Soccer Database

| Approach | $RPS_{avg}$ | Accuracy | Paper |
|---|---|---|---|
| CatBoost+pi-ratings | 0.1925 | 0.5582 | Razali et al. [96]* |
| TabNet+pi-ratings | 0.1956 | 0.5582 | Razali et al [97]* |
| Bookmaker odds | 0.2020 | 0.5194 | Bookmakers [100]* |
| Elo Ordered Logit Model | 0.2035 | 0.5146 | Robberechts & Davis [100]* |
| XGBoost+Berrar ratings | 0.2054 | 0.5194 | Berrar et al. [8]* |
| XGBoost+pi-ratings | 0.2063 | 0.5243 | Hubáček et al. [59] |
| Double Poisson | 0.2082 | 0.4888 | Hubáček et al. [60]** |
| Hybrid Bayesian Network+pi-ratings | 0.2083 | 0.5146 | Constantinou [28] |
| Bradley–Terry, Poisson log-linear hierarchical model, integrated nested Laplace approximation | 0.2087 | 0.5388 | Tsokos et al. [116] |
| Berrar ratings | 0.2101 | 0.4854 | Hubáček et al. [61]* |
| kNN+pi-ratings | 0.2149 | 0.5049 | Berrar et al. [8] |

Studies that were not part of the 2017 Soccer Prediction Challenge are denoted by *.
The ** denotes that this study used a subset of the Open International Soccer Database, from the 2000/2001 to 2005/2006 seasons.

conditions), and team-level characteristics (e.g., average height, attack running rate) could result in improved model performance.

Notably, the three top-performing studies in the 2017 Soccer Prediction Challenge applied machine learning models — namely, gradient-boosted tree models [8, 59] and Bayesian Networks [28] — to feature sets consisting of soccer-specific rating features.

#### 1.2.2.2 Post-challenge studies that have used the Open International Soccer Database

Following the conclusion of the 2017 Soccer Prediction Challenge, several other researchers have utilized the Open International Soccer Database. The approaches and results of these studies in terms of RPS and accuracy are summarized, along with the results from the top 4 2017 Soccer Prediction Challenge studies, in Table 1.2.

Robberechts & Davis [100] used goals- and result-based offensive/defensive models and Elo ratings to predict FIFA World Cup results as well as matches in the Open International Soccer database. Elmiligi & Saad [40] also used the Open International Soccer Database, and proposed a hybrid approach that combined machine learning and statistical models, achieving 0.4660 accuracy and an average RPS of 0.2176. The authors compared the performance when using all matches versus only part of the match history, and when using individual league models versus a model for all leagues. Razali et al. [96] made use of the Open International Soccer Database, comparing the performance of gradient-boosting algorithms including XGBoost,

LightGBM [71], and CatBoost [93] on goals- and result-based Elo ratings [65, 100], as well as pi-ratings. The authors found that applying CatBoost to pi-ratings yielded the best performance, with an average RPS of 0.1925 and accuracy of 0.5582, which outperformed the previous 2017 Soccer Prediction Challenge studies (and other studies that have used the Open International Soccer Database). In another recent study, Razali et al. [97] used a deep learning approach called TabNet [3], a deep neural network designed for tabular data, achieving an average RPS of 0.1956 and accuracy of 0.5582. In a study that only used rating systems and statistical models for results prediction, Hubáček, Šourek & Železný [61] found that Berrar ratings provided the best performance (average RPS: 0.2101, accuracy: 0.4854), followed by Bivariate Poisson, Double Poisson, Double Weibull, and pi-ratings, all of which obtained an average RPS of 0.2103. In earlier work, Hubáček, Šourek & Železný [60] found that on a subset of the Open International Soccer Database (the 2000/2001 to 2005/2006 seasons), the Double Poisson model provided the lowest average RPS of 0.2082 and accuracy of 0.4888 (pi-ratings, however, provided slightly higher accuracy).

Another soccer prediction challenge, using a similar dataset, was held in 2023, however, the results were not available at the time of writing. Please see the appendix of this chapter for a brief description of the 2023 Soccer Prediction Challenge.

## 1.3 Models

In this section, we first discuss the need for a clear model objective. Given this model objective, a guide to selecting a set of candidate models, as well as a list of commonly applied traditional machine learning models in this domain[7] is provided. Then, we discuss model interpretability and its greater importance to some groups relative to others. We then discuss gradient-boosted tree models including XGBoost and CatBoost, which have provided some of the strongest performance in this domain of late. Following this, deep learning models, which have shown great promise in a number of application domains, are discussed in the context of soccer match result prediction.

### 1.3.1 Model Objective

Models can vary in their objectives, and it is important to establish what the objective of a model is at the outset of a match result prediction project or study. The objective of a model might be purely to achieve high predictive performance, e.g., to compete with expert predictions or in competitions. The model's objective may instead/also be used to place bets on match results. If the predictive model is used for betting, there also needs to be consideration of which matches should be bet on, e.g., using criteria such as the Kelly Index [114]. Finally, a model might be used for performance analysis purposes by coaches or analysts. Model interpretability is important for this group so that they are able to identify the most relevant features that are of importance to winning so that past performance can be analyzed and team strategy can be adjusted to increase the chance of winning future matches. Model interpretability is discussed further in subsection 1.3.3.

---

[7] By traditional, we mean excluding gradient-boosted tree and deep learning models, which are relatively recent developments in this domain and are described later in this chapter.

### 1.3.2 Candidate Models

When deciding on a set of candidate ML models for soccer match result prediction, a useful starting point is to thoroughly review the literature and identify models that have performed well in predicting sports match results in general, but also those that have been effective in soccer in particular. The purpose of the model is again relevant when selecting the set of candidate models, as well as who the target audience is since, as mentioned, interpretability is of greater importance to certain groups. On the other hand, if the purpose of the model is purely to achieve the highest performance, black-box models can be sufficient. While some studies have used a single type of model to predict match results, it is more common to conduct a comparative evaluation of the performance of a range of candidate ML models. Furthermore, as well as the 2017 Soccer Prediction Challenge, to confirm model generalizability, some researchers have applied their models to several leagues, e.g., the top 5 European leagues [35, 74, 127], the Greek/Dutch/English leagues [82] and the leagues of 12 different countries [23]. Before recent studies that have used gradient-boosted model trees and deep learning models, the most commonly applied classification ML models for soccer match result prediction included Logistic Regression, Artificial Neural Networks (ANNs), Bayesian Networks, Decision trees, k-NN, Naïve Bayes, Random Forest, and Support Vector Machine (SVM).

### 1.3.3 Model Interpretability

Increasing the interpretability (understandability) of match result prediction models has become an area with increasing research activity recently. As mentioned, with greater interpretability, match result prediction models are of greater use to, e.g., coaches and performance analysts, to identify the most relevant features that are within players' control and can be adapted to increase the chance of winning future matches. Models with greater interpretability can also be used by teams and/or management to adapt their tactical decisions, to identify own and opposition team strengths and weaknesses, decide on appropriate formations, and make player selection and transfer decisions [124]. Moustakidis et al. [84] used explainable ML models using SHapley Additive exPlanations (SHAP) [80] to identify team performance indicator metrics that are of greatest relevance to predicting teams' average scoring performance per season. Another study that used the SHAP method is that of Ren & Susnjak [99]. Yeung, Bunker, & Fujii [124] proposed an interpretable ML model framework for soccer match result prediction that is conceptually similar to GAP Ratings [118], but that predicts match statistics with — rather historical match statistics, which can be cumbersome to engineer — management decision- and player quality-related features. Player quality features were obtained from the EA Sports video game FIFA, a data source that has been used in other studies, e.g., [35, 92]. Random forest feature importance is another appealing approach to interpreting the most relevant model features and has been used in conjunction with a Hybrid Random Forest model by Groll et al. [53]. Of course, some traditional ML models, e.g., decision trees and logistic regression, are also highly interpretable by design and may be useful if they can provide sufficient performance.

### 1.3.4 Ensemble Methods

Ensemble methods in ML can be broadly grouped into boosting methods, e.g., gradient-boosted tree models such as XGBoost and CatBoost, and bagging methods, e.g., Random Forest.

As previously mentioned, at least in the absence of match features other than goals scored, the 2017 Soccer Prediction Challenge and subsequent studies using the Open International Soccer Database have highlighted that gradient-boosted tree models, applied to soccer-specific ratings, are currently able to achieve some of the highest performance in this domain (Table 1.2).

However, on other datasets (some of which include match statistics), Random Forests have been found to be competitive with [5] and even exceed the performance of gradient-boosted tree models [107, 1, 43]. Despite ensemble methods appearing to provide generally better performance than statistical and other traditional ML models, a thorough comparative evaluation of the performance of ensemble methods with deep learning and (deep) neural network models is still required.

Extreme gradient boosting, or XGBoost [26], is one of the most popular gradient-boosted tree methods and has performed well in a variety of ML tasks. The foundation of XGBoost lies in gradient boosting. Gradient boosting, which can be used for both regression and classification, combines several weak learners (e.g., decision trees) into so-called strong learners (gradient-boosted trees). The gradient boosting process sequentially builds simple weak prediction models that each predict the residual error of the preceding model. The weak learners are added to the ensemble and their contributions are determined based on a gradient descent optimization problem, which minimizes an overall loss function representing the difference between the actual results and predictions of the strong learner.

CatBoost [93] has some similarities to XGBoost and is capable of handling both regression and classification tasks by aggregating multiple weak learners. One of the distinctive features of CatBoost is its ability to handle categorical features efficiently, alleviating the need for extensive data preprocessing. CatBoost employs a technique called "ordered target encoding," which involves encoding a categorical feature sequentially while excluding the needs of the target feature. This prevents information leakage (where the model accesses the target information of the current observation), ensuring an unbiased encoding process and preventing overfitting. The generation of match outcome probabilities (e.g., for win/draw/loss) that are well-calibrated is a distinct advantage of CatBoost, given its importance in match result forecasting [121].

Another boosted-tree model that has not been widely investigated, but that may have potential, for soccer match result prediction is the Alternating Decision Tree (ADTree) model [46]. As opposed to XGBoost and CatBoost, ADTree uses AdaBoost [47] rather than gradient boosting, and maintains an interpretable decision tree structure as the final model (see the Appendix for more detail).

### 1.3.5 Deep Learning Models

Deep learning models have been effective in various domains including computer vision, trajectory analysis, and event prediction in sports. Nevertheless, there are still only a relatively small number of published articles related to deep learning for soccer match result prediction.

Danisik, Lacko, & Farkas [35] compared the performance of a Long Short-Term Memory (LSTM) model [56] with classification, numeric prediction, dense approaches, and also against average random guess, bookmaker prediction, and home team win baselines in predicting English Premier League matches. Rahman [94] used deep neural networks and ANNs to predict match results from the 2018 FIFA World Cup. Jain, Tiwari & Sardar [67] used Recurrent Neural Networks and LSTM networks for predicting English Premier League match results, engineering several features related to winning and losing streaks, points, and goal differences. Malamatinos, Vrochidou, & Papakostas [82] used k-NN, LogitBoost, SVM, Random Forest, and CatBoost — as well as convolutional neural networks and transfer learning with encoded tabular data converted to image models — to predict Greek Super League match results, with the best performing model also applied to the English Premier League and the Dutch Eredivisie. The best-performing model was found to be CatBoost, notably outperforming the deep learning convolutional neural network model. Given the time-series nature of soccer match result data, Joseph [69] considered time series-based approaches to predict English Premier League match results, including LSTM and Bayesian methods. As mentioned in subsection 1.2.2.2, Razali et al. [97] recently achieved strong performance (Table 1.2) when applying TabNet [3] — a deep neural network model for tabular data — to the Open International Soccer Database.

Given the relatively small number of studies related to deep learning in soccer result prediction, there is the potential for greater future exploration into the potential of these models, e.g., to investigate whether deep learning models are able to outperform boosted tree and other ensemble models.

## 1.4 Features

This section describes some of the different types of features that can be used in soccer match result prediction models.

Features in sports match result data can generally be divided into different feature subsets. For instance, in basketball, Miljković et al. [83] categorized features into match-related and standings features. Tax and Joustra [113] compared the performance of a combined feature set consisting of betting odds and features derived from publicly available data to a feature set consisting only of betting odds. Hucaljuk and Rakipović [62] compared the performance of a separate expert-selected feature set against their own selected feature set. Feature selection algorithms can be applied to the entire feature set as well as subsets (e.g., feature selection algorithm-selected vs. human-selected features, betting odds included vs. betting odds excluded, rating features, match features, external features, etc.) of the original feature set, and the performance of the candidate models on each feature set can be compared.

In the remainder of this section, first, rating features are discussed, including general-purpose rating systems that are used across a range of sports, e.g., Elo ratings. Domain-specific (soccer-specific) rating systems including pi-ratings, Berrar ratings, and the recently proposed GAP Ratings are then discussed, as are betting odds (which can also be considered a rating for which the calculation is opaque). Then, match, player, and team statistics are examined, as are external features that do not relate to events occurring within matches. Feature selection methods are also briefly discussed.

### 1.4.1 Ratings

The general idea of rating systems is to assign some initial rating for each team and update the ratings over time based on actual match results. Some soccer-specific ratings also consider the goal margin, match venue, and offensive/defensive strengths. There are several use cases for ratings, the first and most obvious of which is to rate/rank teams. In the context of match results prediction, rating systems can, of course, be used as a predictive model by simply predicting the team with the higher rating as the winner. However, using ratings as model features in machine learning models appears to provide better performance (Table 1.2), and for this reason, ratings are described here in the features section of this chapter rather than in the models section.

#### 1.4.1.1 Elo Ratings

The Elo ratings [41] system was originally used in chess but has since been applied in a wide range of sports including soccer. Extensions of Elo have been proposed, e.g., to account for goal margin [65] and home advantage [104].

In the Elo rating system, each team generally starts with a rating of 1500. The system can be decomposed into two steps: the estimation step (E-step) and the update step (U-step). The E-step involves estimating the probability that a team will win a particular match, while the U-step involves the update of the team's rating following a particular match result.

**E-step.** The probability that team $i$ beats team $j$ is estimated by:

$$p_{i,j}(t) = \frac{1}{1 + 10^{\frac{-(R_i(t) - R_j(t))}{400}}} \tag{1.1}$$

Analogously, the probability that team $j$ beats team $i$ is:[8]

$$p_{j,i}(t) = \frac{1}{1 + 10^{\frac{-(R_j(t) - R_i(t))}{400}}} \tag{1.2}$$

Since these are probabilities, they add to one, i.e., $p_{i,j}(t) + p_{j,i}(t) = 1$.

**U-step.** The Elo ratings are updated, e.g., for team $i$, according to:

$$R_i(t+1) = R_i(t) + K[A_i(t) - p_{i,j}(t)]$$

Team $j$'s rating is updated in a similar manner:

$$R_j(t+1) = R_j(t) + K[A_j(t) - p_{j,i}(t)]$$

where $A_i(t)$ takes the value of 1, 0.5, or 0 for a win, draw, and loss, respectively.

**K-factor.** The $K$ term in the above U-step expressions is known as the $K$ factor and is often given a fixed value (e.g., 32 or 20). If $K$ is too large, the Elo rating will vary excessively from match to match; however, if $K$ is too small, the rating of a team will not change rapidly enough when it improves its performance [65].

There are different approaches to selecting the value of $K$: it can be a fixed value, a tunable parameter, or a function. Hvattum & Arntzen [65] proposed a goal-based Elo rating system that allows $K$ to depend on the goal difference. In particular, $K = K_0(1 + \delta)^\lambda$, where $\delta$ is the absolute goal margin, and $K_0$ and $\lambda$ are parameters.

Other approaches to selecting the $K$-factor are described online, e.g., on elorat-ings.net and footballdatabase.com (see the Appendix for further details).

**Elo Ratings with Home Advantage.** Ryall & Bedford [104] extended Elo ratings to incorporate home advantage (albeit in Australian Rules Football, but it can equally be applied in soccer) by simply adding a home advantage term to the rating difference such that the estimated probability of team $i$ beating team $j$ from Equation 1.1 becomes

$$p_{i,j}(t) = \frac{1}{1 + 10^{\frac{-(R_i(t) - R_j(t)) + H_{i,j}}{400}}} \tag{1.3}$$

---

[8] The value of 400 was specified by Elo [41] to be twice the standard deviation of chess player ratings. The values of 10 and 400 merely serve to set a scale for the ratings and are generally used when the initial ratings are set to 1500 [65].

where $H_{i,j}$ is the magnitude of home advantage that team $i$ has over $j$. Note that in this setup, $H_{i,j}$ is a tunable parameter that does not vary by match, $t$.

**Connection between Elo and Bradley-Terry.** As mentioned earlier, the Bradley-Terry paired comparisons model [11], which estimates the win probability in a match using the strengths/abilities of each team, is the foundation of Elo ratings. In the Bradley-Terry model, the probability of team $i$ beating team $j$ is given by:

$$p(i \text{ beats } j) = \frac{s_i}{s_i + s_j}$$

where $s_i$ and $s_j$ denote the strength (ability) of teams $i$ and $j$, respectively. The strength parameters are commonly estimated using maximum likelihood, e.g., based on observed win counts between the two teams. The Elo rating of a team can actually be expressed directly as a function of its Bradley-Terry strength [33]. In particular, the Elo rating of team $i$ can be expressed as $R_i = 400 log_{10}(s_i)$. Rearranging this expression enables one to calculate the Bradley-Terry strength of team $i$ from its Elo rating: $s_i = 10^{\frac{R_i}{400}}$.

The Bradley-Terry model has been shown to be outperformed in soccer match result prediction by Poisson/Weibull-based models [61]. Furthermore, when using the ratings themselves for prediction, pi-ratings have been found to generally outperform Elo ratings.

### 1.4.1.2 Pi-ratings

In the pi-ratings system [30], each team is assigned two ratings corresponding to their home and away strengths, with their overall rating the average of their home and away ratings:

$$R_i = \frac{R_{i,H} + R_{i,A}}{2}$$

Then, for each match, the expected goal difference is calculated based on the home rating of the home team and the away rating of the away team. The actual match outcome is compared with the expected score and if a team performs better than expected, its rating is increased based on the difference between actual and expected outcomes, as well as the learning rates (which are model parameters). The home and away ratings of both teams are updated after a match but using separate learning rates. Each team starts with an initial pi-rating of 0, and the pi-rating represents the rating of the average team relative to other teams. Given a match between teams $\alpha$ and $\beta$, the ratings are updated according to:

$$\hat{R}_{\alpha,H} = R + \lambda \psi_H(e)$$

$$\hat{R}_{\alpha,A} = R_{\alpha,A} + \gamma(\hat{R}_{\alpha,H} - R_{\alpha,H})$$

$$\hat{R}_{\beta,H} = R_{\beta,H} + \lambda \psi_A(e)$$

$$\hat{R}_{\alpha,H} = R_{\alpha,H} + \gamma(\hat{R}_{\alpha,A} - R_{\alpha,A})$$

where $e$ is the error between the expected and actual goal difference. The expected goal difference for a team against the average opponent at ground $G$, which is either $H$ (home) or $A$ (away), is calculated as:

$$\hat{g}_{DG} = b^{\frac{|R_{iG}|}{c}} - 1$$

where $b = 10$ and $\hat{g}_{DG}$ is the expected goal difference for team $i$ against the average opponent when playing at ground $G$ (either $H$ or $A$). Team ratings can potentially be negative, in which case the expected outcome is $-\hat{g}_{DG}$. The error, $e$, between the expected and actual goal difference is then given by:

$$e = |g_D - \hat{g}_D|$$

where $g_D = g_{DH}$ - $g_{DA}$ and $\hat{g}_D = \hat{g}_{DH}$ - $\hat{g}_{DA}$ To dampen the influence of large goal margins when updating ratings, a reward/penalty function is specified to be a function of the goal difference error:

$$\psi(e) = c \times log_{10}(1 + e)$$

where $c = 3$. The probability distribution over potential match outcomes is computed using an ordered logit model.

### 1.4.1.3 Berrar Ratings

In the Berrar ratings system [8], a logistic function is used to predict the expected number of goals scored using the teams' offensive and defensive strengths. In particular, the expected goals scored by the home and away teams are respectively given by:

$$\hat{G}_H = \frac{\alpha_H}{1 + exp(-\beta_H(o_H - d_A) - \gamma_H}$$

and

$$\hat{G}_A = \frac{\alpha_A}{1 + exp(-\beta_A(o_A - d_H) - \gamma_A}$$

where $o_H$ and $d_H$ and $o_A$ and $d_A$ denote the offensive and defensive ratings of the home and away teams, and $\alpha_H$ and $\alpha_A$ represent the maximum possible expected goals that can be scored in a match by the home and away teams. The $\beta$ parameters determine how steep the logistic function is, and $\gamma$ determines the threshold/bias.

The ratings for the home team are updated according to:

$$o_H = o_H + \omega_{o_H}(G_H - \hat{G}_H)$$

$$d_H = d_H + \omega_{d_h}(G_A - \hat{G}_A)$$

And similarly, for the away team:

$$o_A = o_A + \omega_{o_A}(G_A - \hat{G}_A)$$

$$d_A = d_A + \omega_{d_A}(G_H - \hat{G}_H)$$

#### 1.4.1.4 GAP Ratings

Since soccer is a low-scoring sport and goals are rare, more recently, Generalized Attacking Performance (GAP) Ratings [118] were proposed, which build on pi-ratings but can also instead predict non-rare match statistics (e.g., shot attempts), not only goals scored. By predicting non-rare match statistics and excluding some of the early and latter rounds of seasons, GAP ratings were shown to be more informative (in terms of the Akaike Information Criterion) for soccer match result forecasting [120].

GAP ratings [118, 120] are inspired by pi-ratings [30] and are adjusted such that they can be applied to predict match statistics in advance. The GAP rating requires match statistics as input and can be used to predict goals, shots on and off target, and corners for both teams. The inputs, $S_H$ and $S_A$, denote the match statistics for the home team ($H$) and away team ($A$), respectively. There are four ratings for each team $i$: $H_i^a, H_i^d, A_i^a, A_i^d$, which denote team $i$'s home attack, home defence, away attack, and away defence ratings, respectively. Following a match between home team $i$ and away team $j$, the GAP ratings for home team $i$ are updated as follows:

$$H_i^a = max(H_i^a + \lambda\phi_1(S_H - \frac{H_i^a + A_j^d}{2}), 0)$$

$$A_i^a = max(A_i^a + \lambda(1 - \phi_1)(S_H - \frac{H_i^a + A_j^d}{2}), 0)$$

$$H_i^d = max(H_i^d + \lambda\phi_1(S_A - \frac{A_j^a + H_i^d}{2}), 0)$$

$$A_i^d = max(A_i^d + \lambda(1 - \phi_1)(S_A - \frac{A_j^a + H_i^d}{2}), 0)$$

The GAP ratings for away team $j$ are updated as follows:

$$A_j^a = max(A_j^a + \lambda\phi_2(S_A - \frac{A_j^a + H_i^d}{2}), 0)$$

$$H_j^a = max(H_j^a + \lambda(1 - \phi_2)(S_A - \frac{A_j^a + H_i^d}{2}), 0)$$

$$A_j^d = max(A_j^d + \lambda\phi_2(S_H - \frac{H_i^a + A_j^d}{2}), 0)$$

$$H_j^d = max(H_j^d + \lambda(1 - \phi_2)(S_H - \frac{H_i^a + A_j^d}{2}), 0)$$

where $\lambda > 0$ is a parameter that weights the effect of the latest game on the rating, and $\phi_1 \in [0, 1]$ and $\phi_2 \in [0, 1]$ weights the effect of a home game on the away rating and an away game on the home rating, respectively.

Using goals as an example, $H_i^a$ indicates the number of goals home team $i$ will score and $A_j^d$ indicates the number of goals the away team $j$ will concede. The average of $H_i^a$ and $A_j^d$ represents the expected goals home team $i$ should score and away team $j$ should concede. Ratings are updated using the weights and the differences between actual and expected results. The GAP rating can be interpreted as the performance of a team compared to an average team in the league. Better teams will have higher attacking and lower defensive ratings. The estimates of $S_H$ and $S_A$, which are denoted by $\widehat{S_H}$ and $\widehat{S_A}$, respectively, are given by:

$$\widehat{S_H} = \frac{H_i^a + A_j^d}{2}$$

$$\widehat{S_A} = \frac{A_j^a + H_i^d}{2}$$

To select the $\lambda$, $\phi_1$, and $\phi_2$ parameters, the least-squares method is applied. The cost function is given by:

$$Cost = \sum_{i=1}^{N} |S_H - \widehat{S_H}| + |S_A - \widehat{S_A}|$$

where $N$ denotes the number of matches. All GAP ratings are initialized to zero. Parameter selection is repeated at the start of each season using all data. However, the first year of data is only used for parameter selection, and the first six games and final six games played by the home team in each season are ignored due to various factors that could increase the unpredictability of these matches. Also, the relegated teams' ratings are calculated as the average of the promoted teams' ratings, and similarly, the promoted teams' ratings are calculated as the average of the relegated teams' ratings.

### 1.4.1.5 Betting Odds

As mentioned, betting odds can be used as a baseline with which to model results (subsection 1.4.1.5), but they can also be considered a type of rating [122]. Therefore, like other ratings, odds can potentially be used as a model feature. Indeed, betting odds have been used as model features in many studies [106, 86, 113, 44], sometimes as the sole model feature. Since bookmakers establish betting odds, unlike the rating systems described above, the way in which odds are determined is opaque. However, it is reasonable to assume that bookmaker companies use models supplemented with observed betting volumes on each outcome to capture the "wisdom of crowds" [111, 17] in order to set appropriate odds for matches. Betting odds also differ from other ratings that generally only incorporate historical match result information, since odds also account for other factors such as market sentiment, player availability, and expert knowledge.

Deciding whether betting odds should be included as a model feature depends on the objective of the model. For instance, if a model is being used for betting purposes in an attempt to "beat the house," betting odds should not be included as a model feature [58]. As a baseline, betting odds have proven difficult to beat. For example, despite using a state-of-the-art gradient boosting model along with a sophisticated engineered set of features, Baboota & Kaur [5] were unable to outperform predictions based on bookmaker odds. Bookmaker predictions also outperformed the top-performing submissions from the 2017 Soccer Prediction Challenge (odds outperformed all competing models in [100], and this result is also shown in Table 1.2, along with the next-best result from that study, an Elo ordered logit model). On the other hand, if the objective is purely maximizing predictive performance, given the information inherent in betting markets, including betting odds model features may make sense.

Some studies have attempted to exploit bookmaker odds while using machine learning models, e.g., through arbitrage [74] and modern portfolio theory methods [58], including Talattinis et al. [112], who utilized the Sharpe Ratio. Many betting sites provide betting odds data for future and historical matches (Table 1.1). When sourcing betting odds from such websites, to create a model feature, it is often useful to convert the raw decimal odds into probabilities by taking their reciprocal:

$$\text{Probability} = \frac{1}{\text{Decimal Odds}} \qquad (1.4)$$

The sum of the odds generally exceeds one due to the presence of a built-in bookmaker margin, which is sometimes referred to as the "over-round" [45, 125]. To ensure that the odds add to one and obtain the normalized odds-implied win probabilities, one simply divides the original odds-implied probabilities by the normalization factor, which is the sum of the probabilities obtained for each match outcome from equation 1.4. That is,

$$\text{Probability Normalized} = P_{norm} = \frac{\text{Probability}}{\text{Normalization Factor}}$$

For example, suppose that a match has only two possible outcomes and that the decimal bookmaker odds are such that team A is paying 1.34 to win and team B is paying 3.02. Taking the reciprocal as per equation 1.4, we obtain 0.746 and 0.331 for teams A and B, respectively. Then, 1.077 is the normalization factor ($0.746 + 0.331 = 1.077$). To obtain the odds-implied win probabilities, we have $P_{norm}(\text{team A win}) = 0.746/1.077 = 0.693$ and $P_{norm}(\text{team B win}) = 0.331/1.077 = 0.307$, so now $P_{norm}(\text{team A win}) + P_{norm}(\text{team B win}) = 0.693 + 0.307 = 1$.

Of course, betting on the match outcome is not the only option for bettors: it is possible to bet on a specific event occurring within a match, who will score the first goal, and so on. Betting on the over-under — whether a team scores over or under 2.5 goals in a match — is one such option. Wheatcroft [118] found that in the over/under 2.5 goals market, shots made and corners were better for probabilistic forecasting than actual goals.

### 1.4.2 Match Features

**Simple Match Statistics & Performance Indicator Metrics:** As has been mentioned in previous sections, match features (also known as in-play, in-match, or in-game features) are features that relate to events that occur within matches. Match features are often performance indicator metrics [63], as they are referred to in the field of sports performance analysis, a sub-discipline of sports science. For instance, raw statistics such as passes, tackles, shot attempts, etc., can be aggregated into performance indicator metrics at the player, offensive/defensive unit, or team levels. The survey paper of Bunker & Susnjak [18] highlighted the potential benefits of greater future collaboration between researchers from sports performance analysis and machine learning for engineering relevant match features for machine learning models.

Of course, match features that are performance indicators are not known in their entirety until the match has concluded. Therefore, for match result forecasting, preprocessing is required, by aggregating (e.g., averaging) match features across a certain number of historical matches. There is no set number of matches with which to aggregate/average over since it will depend on the characteristics of the dataset. For example, Buursma [22] found that aggregating features over the past 20 matches provided the best performance, while Berrar et al. [8] found that using the past nine matches provided the best performance in building a set of what the authors referred to as "recency features" for the 2017 Soccer Prediction Challenge. Simple aggregation (e.g., averaging or summing) of features over a certain number of past matches does not, however, account for the fact that recent games are of greater relevance than matches that were played a long time ago. Exponential time weighting [37, 78, 60], which is used in the weighted likelihood function that is maximized in Double Poisson models, could be used as an alternative to simple aggregation over historical matches.

Matches can be summarized in the form of simple statistics, some of which are performance indicators and some are not. Goals are the most commonly used match statistic since goals are also used to derive the match result, and are therefore used in essentially every study on match result forecasting. Shots on target, shots off target, and corners are other commonly used match statistics. Other match statistics in soccer that can also be considered performance indicators include blocks, clearances, crosses, dribbles, interceptions, passes, possessions, saves, and tackles. Discipline-related match statistics include yellow and red cards, fouls, and offsides. As mentioned, match statistics prediction is an important part of the GAP Ratings method [118], which focuses on predicting non-rare events such as shot attempts (since actual goals are rare).

**On-the-Ball Event data**: Different vendors and data providers, e.g., StatsBomb and Stats Perform/Opta, have varying data formats and definitions that are used to describe and annotate events that occur within soccer matches. Decroos et al. [36] proposed a Soccer Player Action Description Language (SPADL) to standardize the event data format across vendors. The authors also proposed a method called Valuing Actions by Estimating Probabilities (VAEP), which uses event log data that has been converted into SPADL format, to value on-the-ball actions. VAEP can also be used to compute player ratings by aggregating each of their on-the-ball actions. Event data commonly includes the start and end times and locations of events, the player who performed the action and the team to which they belong, the action type (e.g., pass, cross, throw in, etc.), the body part used in the action, and whether the action was successful or unsuccessful. While event data focuses event on on-the-ball rather than off-ball events, supplementing on-the-ball event features with tracking data-derived features can address this.[9] Event data can provide more in-depth information than simple match statistics and has the potential to be further explored for engineering informative features for soccer match result prediction ML models.

**Spatiotemporal tracking data**: Spatiotemporal tracking data is generally sourced from optical systems or wearable (GPS) tracking devices that record the locations of players and the ball. Another way in which spatiotemporal data can be obtained is from computer vision, machine learning, or deep learning methods applied to match video footage. As mentioned, it is valuable for event data and spatiotemporal data to be combined to be able to perform more holistic performance analyses, however, this is challenging – both in terms of cost but also technically — because it generally requires collaboration between researchers from disparate disciplines, i.e., computer/data scientists and sports scientists [49]. Furthermore, professional teams are generally unwilling to share data publicly, and obtaining it from professional data providers comes at significant cost. Rein & Memmert [98] also highlighted the value of greater collaboration between sports scientists and computer scientists in extracting tactical insights from spatiotemporal tracking data in soccer, which has traditionally been used for monitoring physiological demands. Toda et al. [115] proposed a method called Valuing Defense by Estimating Probabilities (VDEP), an adjustment of VAEP in which the defensive performance of a team is evaluated using on-the-ball event and tracking data. The generated metrics focus on actions that lead to penetration of the penalty area or a shot attempt (rather than a goal, which is rarer). The study showed that this resulted in higher predictability than VAEP. Deriving features from spatiotemporal tracking data also appears to have the potential for further research in match results prediction ML models in soccer. As mentioned, spatiotemporal tracking data has traditionally been used to assess physiological demands on athletes, and Tümer et al. [117] aimed to predict Turkish Super League rankings by applying three ML models – ANN, radial basis function, and linear regression — to physical and technical features (e.g., related to physical demands).

---

[9] F. Goes (2021), The power of combining tracking and event data, https://www.scisports.com/the-power-of-combining-tracking-and-event-data/

**Expected Goals**: Expected Goals (xG) refers to the expected number of goals a team should score in a game. For every shot a team takes in a match, the probability of a goal can be calculated using xG models, of which there are many types. Then, by aggregating all of these probabilities, one can obtain the xG for the match. To build an xG model, shot-related features are considered. For example, according to the Bundesliga,[10] the distance and angle to the goal, whether it is a pass or a header, and the type of pass are used. For the algorithm, any type of regression model, e.g., logistic regression or neural network, can be utilized. The target variable takes the value of 1 for a goal and 0 for no goal. Wheatcroft [120] showed that using goals could not provide statistically significant information. Replacing goals with expected goals (xG) as the target variable is a commonly suggested approach for further research because expected goals are not as rare.

### 1.4.3 Player Statistics

**Player characteristics and ratings.** The incorporation of player-level attributes or ratings has been a popular line of enquiry in recent times. For example, Stübinger, Mangold, & Knoll [108] incorporated player and match attributes in an ensemble learning approach in a simulation on the top 5 European leagues. Chen [25] used FIFA video game player ratings to predict Spanish La Liga soccer results using three common ML models: ANN, Random Forest, and SVM. Another study that utilized FIFA player ratings is that of Danisik, Lacko, & Farkas [35], whose deep learning approach using LSTM has been discussed previously in this chapter. Arntzen & Hvattum [4] compared the performance of an ordered logistic regression model and a competing risk model when applied to adapted plus-minus player ratings — which compare the performance of a team, e.g., in terms of goals, when a particular player is playing versus when they are not [64]) — with when the two models were applied to team Elo ratings. The authors found no difference in performance between the ordered logit and competing risk models for match result prediction. Furthermore, they found that team ratings and player ratings had similar performance, but using a combined feature set with both player ratings and team ratings provided better performance than when utilizing only player ratings or only team ratings. Team and player ratings from EA Sports' FIFA are available via websites such as fifaindex.com (Table 1.1). Other video games such as Football Manager (Sports Interactive/Sega) also contain player ratings that are arguably better and could potentially be used in future studies.

**Player Form**: Otting and Groll [88] applied Hidden Markov models to investigate the existence of the "Hot Shoe" effect (known as the "Hot Hand" effect in basketball), and provided evidence that such an effect does exist. If such an effect does exist, it is reasonable to assume that players in the "hot" state are more likely to perform better than those in the "cold" state, which could provide more information for prediction.

---

[10] Bundesliga (2019), xG stats explained: the science behind Sportec Solutions' Expected goals model, https://www.bundesliga.com/en/bundesliga/news/expected-goals-xg-model-what-is-it-and-why-is-it-useful-sportec-solutions-3177

### 1.4.4 Team Statistics

**Team Ratings.** In subsection 1.4.1, team ratings based on historical match results in terms of goals scored were discussed. However, there are other ways to obtain or compute team ratings. The first approach is to obtain team ratings from video games such as FIFA or Football Manager and directly use these as features in a predictive model [5, 124]. Aside from video games, another potential source of team ratings is the Union of European Football Associations (UEFA), which provides team ratings based on results in UEFA club competitions over the past five years [109]. Player ratings can be obtained from video games such as FIFA and then aggregated to engineer role- or team-level ratings [24]. Carpita, Ciavolino and Pasca [24] used 33 player-related features from the FIFA video game to construct seven player-level performance indicators, which were then combined into performance indicators based on role (forward, midfielder, defender and goalkeeper). A binomial logistic regression model was then applied to differences between the role-based performance indicators of opposing teams to analyze the degree to which these indicators affect the probability of winning. Alternatively, player ratings can be computed using VAEP or plus-minus ratings using event log data or historical match result data, respectively, and then aggregated into a team rating.

Some studies have incorporated or compared both player and team ratings. For instance, Pipatchatchawal & Phimoltares [92] incorporated both player and team FIFA video game ratings, while Arntzen & Hvattum [4], as previously mentioned, compared the use of team Elo ratings and player plus-minus ratings.

**Streak**: Streak is a similar concept to the hot shoe phenomenon; however, it generally refers to match result patterns rather than individual outcomes such as successful shots on goal. Baboota & Kaur [5] computed the team streak as of a team's $j$th game as:

$$Streak(j) = \sum_{i=j-k}^{j-1} Score_i/3k$$

where $Score_i$ indicates the goals scored in game $i$, $Score_i \in \{0, 1, 3\}$ and $k$ is a tunable hyperparameter indicating how many historical games are considered. To account for the fact that more recent matches are of greater relevance, the authors also engineered a time-weighted streak feature, which for the $j$th game, is given by:

$$Weighted\ Streak(j) = \sum_{i=j-k}^{j-1} 2 * \frac{i - (j - k - 1)Score_i}{3k(k + 1)}$$

The experimental results of the study showed that $k = 6$ provided the best performance for the various models.

**Form**: The streak only considers a single team; thus, to consider the opponent teams, Baboota & Kaur [5] also engineered a form feature. The form of each team is set to an initial value of one and is updated after each game. The forms of teams $A$ and $B$, for the $j$th game between these two teams, if team $A$ wins, are given by:

$$Form_j^A = Form_{j-1}^A + \alpha Form_{j-1}^B$$

$$Form_j^B = Form_{j-1}^B - \alpha Form_{j-1}^A$$

If the match is a draw, the forms are given by:

$$Form_j^A = Form_{j-1}^A - \alpha (Form_{j-1}^A - Form_{j-1}^B)$$

$$Form_j^B = Form_{j-1}^B - \alpha (Form_{j-1}^B - Form_{j-1}^A)$$

The authors found that the hyperparameter $\alpha = 0.33$ provided the best performance.

**Inter-player Chemistry**: Bransen and Van Haaren [12] suggested that the chemistry between players within the same team is more important than the past performance of a player in another team. Two offensive and defensive chemistry metrics were proposed based on VAEP, which considered the interaction between two players, to identify the squad with the best chemistry. Moreover, two CatBoost models were applied to player statistics features for a pair of players, with the two chemistry metrics as the respective target variables, in order to predict the chemistry of a specific player with players in other teams (which is potentially useful for scouting or transfers).

**Passing Networks:** Some studies have used social or passing network analysis to predict match results in soccer. Using passing distribution data, Cho, Yoon and Lee [27] used social network analysis to construct actual and predicted network indicators that reflect team performance, in conjunction with gradient boosting for match result prediction. Ievoli, Palazzo, & Ragozini [66] considered whether passing networks have a significant effect on match results, applying four ML models to passing network-derived indicators and an on-field feature set to predict 2016–2017 UEFA Champions League group stage results. The authors found that some network-derived variables were connected to the level of offensive actions, and could improve the explanatory power of models beyond the models that included only on-field features.

### 1.4.5 External Features

External features are external to the match itself, i.e., these features are not derived from on-field events. These may include travel, player availability, match venue, match officials, weather, and so on. One study that incorporated weather as a model feature is that of Palinggi [89], who applied an SVM model to weather- and match-related features and achieved around 50% accuracy. Other potential external model features include the average age of a team's players and coach [52], the number of players who are international representatives [70], as well as club market values, transfer budgets, and operational costs [109]. Unlike match features, external features are generally already known prior to a match and can, therefore, often be used directly as model features, without the type of preprocessing that is required for match features through aggregation over historical matches. In terms of match venue having an effect, the home advantage phenomenon is a well-known effect in which the home team tends to outperform the away team, on average (the COVID-19 pandemic period, where some matches were played in front of empty stadiums, being a possible exception [77]).

Another potential source of external feature data is social media. Wunderlich & Memmert [123] applied sentiment analysis methods to tweets posted on Twitter during over 400 English Premier League matches, as well as those posted in the periods shortly before and after goals, to extract information (e.g., word clouds and other relevant features) to predict the total number of goals scored in a match and perform in-play forecasting. Their results suggested that in-play goal forecasting is very challenging, in-play information did not improve predictive accuracy and its predictive value is small in comparison to pre-match information, and that — perhaps due to overly high pre-match expectations from fans — fan sentiment on Twitter decreases during a match. Kinalioğlu & Kuş [72] proposed a hybrid clustering and classification method and applied it to data from 6,396 European league matches including — as well as team and player statistics — fan opinions from social media. Another study that used the Twitter posts of fans to predict soccer match outcomes is that of Kampakis & Adamides [70].

### 1.4.6 Feature Selection Methods

Feature selection techniques can be broadly grouped into three categories: filter, wrapper, and embedded methods [126].

The Random Forest algorithm [13] has embedded tree-based feature selection, in which feature importance is evaluated during the training process [79]. This embedded feature selection means that despite being a bagging method, Random Forest is a model with potentially high interpretability. In the sklearn package in Python, the SelectFromModel object can be used in conjunction with the RandomForestClassifier to automatically select features.

Filter feature selection methods generally rank features based on, e.g., the chi-squared, information gain (ratio), or correlation between each feature and the target variable. A disadvantage of using filter methods is that a cut-off needs to be arbitrarily selected to determine how many features to include in a model.

Other feature selection methods, e.g., CFS subset [55] and Relief/ReliefF subset feature selection methods [73, 75], consider the correlation or interaction among features as well as those between each feature and the target variable. Sequential forward selection is another possible approach to selecting features (e.g., [82]).

As previously mentioned, the performance of models applied to features selected by feature selection techniques can be compared with human expert-selected features (e.g., [62]).

## 1.5 Evaluation Methods

Selecting an appropriate evaluation metric is important when evaluating machine learning models. It is also necessary to clearly define the target variable that is to be predicted.

In sports with only two possible outcomes, assuming the league is competitive, class imbalance is generally not an issue since home team wins should only slightly outweigh away team wins due to the home advantage effect. In this case, classification accuracy can be an appropriate evaluation metric.

In soccer, which has three possible outcomes (with a draw much less common than the other two outcomes), accuracy has continued to be commonly used to evaluate classification ML models. However, more recently, scoring rules such as the Ranked Probability Score (RPS) have become prevalent as evaluation metrics, in part due to its use in the 2017 Soccer Prediction Challenge.

### 1.5.1 Problem Setup & Target Variable Definition

Soccer match result prediction problems are commonly formulated as a three-class classification problem with a target variable defined with three discrete values (e.g., win/draw/loss),[11] as a numeric prediction problem predicting a goal margin target variable,[12] or by predicting the number of goals scored by each team and thus the match result. Since soccer is a low-scoring sport, numeric prediction of the goal difference is challenging compared to sports with match result margins that are larger in magnitude (e.g., basketball or rugby). In machine learning for soccer result prediction, classification is the most common approach, while statistical models commonly attempt to model and predict the number of goals scored by each team. However, given the 2023 Soccer Prediction Challenge (described further in the Appendix) involved —- as one of the tasks — predicting the number of goals scored by each team, more future studies may begin to consider this problem setup.

Other studies have sought to predict other target variables. For example, predicting the "over-under": whether more than 2.5 goals are scored in total in a match [89], or forecasting match statistics such as shot attempts or shots on target [118, 124]. Using a Bayesian approach for prediction, Robberechts, Van Haaren, & Davis [101] developed an in-game win probability model. Wunderlich & Memmert [123] considered both in-play forecasting and predicting the total goals scored in a match. Predicting whether both teams will score during a match has also been considered (e.g., by [32]). Others have attempted to predict the final team rankings in a league [117], or whether a team earns a competition point [43]. The per-season average scoring performance of teams has also been considered as a target variable to be predicted [84].

---

[11] In classification, some researchers, e.g., [72], excluded draws or merged draws into another class so as to create a binary problem, which naturally gives better performance compared to three-class classification models given the difficulty of predicting draws.

[12] Another approach proposed is to convert win/draw/loss to 1/0.5/0 and use a numeric prediction model. This was an approach followed by [35], who used LSTM regression and found that it outperformed the LSTM classifier.

## 1.5.2 Baselines

Since datasets in this domain often cover different leagues and seasons, and contain different variables from which model features can be derived, it is generally difficult to compare results across studies unless the studies being compared utilised a common dataset.

Baselines — including simple rule-based benchmarks, betting odds-derived predictions, and expert predictions — are another way in which researchers can evaluate model performance:

- **Betting Odds:** Betting odds are somewhat unique in that they can be included as a model feature but can also act as a benchmark with which to evaluate model performance. It is straightforward to convert decimal odds into outcome probabilities by taking their reciprocal and normalizing such that the probabilities add up to one (subsection 1.4.1.5). One can then predict the match outcome by simply predicting the outcome that has the highest probability. The odds of several bookmakers can also be combined to create a bookmaker consensus model [76]. Alternatively, provided the model is not being used for betting strategies to "beat the house," these probabilities can potentially be used as model features. The outcome with the highest probability will, of course, almost always be a non-draw outcome.
- **Simple rule-based benchmarks:** A rule that always predicts the majority class (also known as the Zero Rule algorithm) can be used as a baseline. The Zero Rule algorithm (ZeroR), which generally predicts a home team win because of the home team advantage phenomenon, is available in machine learning tools such as WEKA, as are other simple rule-based algorithms such as OneR [57], which can also be used as a benchmark. Random guesses are easily implementable and are also often used as a benchmark, e.g., [35].
- **Expert predictions:** Comparing model predictions with expert predictions, e.g., those of media or former players, is another approach to benchmarking; however, access to experts can be a challenge.[13] Butler, D., Butler, R., & Eakins [21] compared the accuracy of experts and laypeople over three English Premier League seasons and found that former professional soccer players in particular were superior in predicting match results.

### 1.5.3 Scoring Rules

Machine learning models often generate probabilities associated with each class label, and the performance of models — based on how closely these probabilities match the actual (true) class labels — can then be evaluated. Some machine learning models require that match instances are labeled with target variables encoded with a vector with three elements, e.g., in soccer, (1,0,0), (0,1,0) and (0,0,1) can denote win, draw, and loss outcomes, respectively. By representing labels as vectors in this manner, scoring rules such as the Ranked Probability or Brier scores can be utilized, in which a value of 0 indicates a perfect prediction and 1 represents a completely incorrect prediction (thus, lower scores are preferred).

#### 1.5.3.1 Accuracy

Classification accuracy is a widely used evaluation metric and has also been used in many studies in ML for soccer match result prediction. Accuracy is computed by taking the total number of instances (matches) correctly classified by the model (where the predicted value equals the true value) and dividing this by the total number of instances, i.e.,

$$Accuracy = \frac{C}{N}$$

where $C$ denotes the number of instances correctly classified by the model and $N$ is the total number of instances.

#### 1.5.3.2 Brier Score (BS)

The original Brier Score [14] can be applied for multi-class prediction. The lower the BS, the smaller the prediction error. The original Brier Score is given by:

$$\text{BS Original} = \frac{1}{N} \sum_{n=1}^{N} \sum_{i=1}^{R} (\widehat{y_{in}} - y_{in})^2$$

---

[13] Rather than getting experts to provide their match result predictions, other ways of incorporating expert opinion is into the model itself [68] or by engaging experts to select model features [62]. For more details on incorporating expert knowledge, please refer to the Appendix.

The most common version of the Brier Score is for a binary prediction, in which case it is equivalent to the Mean Square Error (MSE) and is given by:

$$BS = \frac{1}{N} \sum_{n=1}^{N} (\widehat{y_n} - y_n)^2$$

The BS was deemed inappropriate in the 2017 Soccer Prediction Challenge since it only measures the difference between the predicted and actual scores but does not account for the ordinal nature of the win/draw/loss target variable.

### 1.5.3.3 Ranked Probability Score (RPS)

The Ranked Probability Score (RPS) [42] does account for the ordinal nature of the three match outcomes in soccer [29] and was thus deemed more appropriate for model evaluation in the 2017 Soccer Prediction Challenge [39]. The RPS is given by

$$RPS = \frac{1}{r-1} \sum_{i=1}^{r-1} \left( \sum_{j=1}^{i} (p_j - a_j) \right)^2, \qquad (1.5)$$

where $r$ denotes the number of potential match outcomes (e.g., $r = 3$ if there are three possible outcomes: home win, draw, and away win). The RPS value always lies within the interval [0, 1], with an RPS closer to 0 representing a better prediction. The RPS can be averaged across multiple instances as follows:

$$\text{RPS}_{\text{avg}} = \frac{1}{N} \sum_{i=1}^{N} \text{RPS}_i. \qquad (1.6)$$

where $i$ denotes the $i$-th instance in the dataset, $RPS_i$ is the $RPS$ value corresponding to that instance, and $N$ is the total number of instances.

Participants in the 2017 Soccer Prediction Challenge aimed to minimize the average RPS across the $N = 206$ matches in the challenge prediction set (containing matches that participants aimed to predict):

$$\text{RPS}_{\text{avg}} = \frac{1}{206} \sum_{i=1}^{N=206} \text{RPS}_i. \qquad (1.7)$$

### 1.5.3.4 Ignorance score (IGN)/Log Loss

The Ignorance score (IGN) was proposed by Good [50], with its foundation lying in information theory. In short, the IGN penalizes predictions with larger logarithmic errors and is given by:

$$IGN = \frac{1}{N} \sum_{n=1}^{N} -(y log_2(p) + (1 - y)log_2(1 - p))$$

Where $y \in \{0, 1\}$ and $p = P(y = 1)$. Its value falls in the range $IGN \in [0, \infty)$, with a lower score representing better model performance. If the log base is changed to base $e$ (multiplied by a constant scalar), the IGN is equivalent to log loss.

Based on its properties, IGN has been suggested as being more appropriate than RPS for evaluating the probabilistic forecasts of soccer match result prediction models [119].

### 1.5.3.5 Root Mean Squared Error (RMSE)

The Root Mean Squared Error (RMSE) evaluates numeric prediction models, e.g., that predict the goals scored or goals margin. The RMSE is calculated by first computing the error, the predicted value minus the actual value, and then taking the square root across all instances:

$$RMSE = \sqrt{\frac{1}{N} \sum_{i=1}^{N} (y_i - \hat{y}_i)^2}$$

where $N$ denotes the number of instances, $y_i$ is the actual observed value of the $i$-th instance, and $\hat{y}_i$ is the predicted value $i$-th instance.

### 1.5.3.6 Scoring rules selection and properties

For numeric prediction problems, RMSE is the most commonly applied scoring rule, while for outcome prediction, the properties of the scoring rule need to be considered. The following three properties are the most commonly described:

- **Proper**: A scoring rule is considered proper when — given inputs $S(P, x)$ with a predicted distribution $P$, an actual distribution $Q$, an outcome $x \sim Q$, and scoring rule $S$ — the rule returns a real value $\mathbb{R}$, which we aim to maximize. The expected score is given by $E[S(P, x)] = S(P, Q)$.
  A scoring rule is *strictly proper* if $S(Q, Q) \geq S(P, Q)$ if and only if $Q = P$, which means that the true distribution is given the lowest (most favorable) score and any other distribution receives a higher (less favorable) score.
  On the other hand, a scoring rule is *proper* if $S(Q, Q) \geq S(P, Q)$ for all distributions $Q$ and $P$. A more theoretical definition can be found in [48].
- **Locality**: A scoring rule is considered local if it only considers the prediction of the observed (target) class. On the other hand, if the scoring rule considers unobserved classes, it is classified as non-local [119].
- **Sensitivity to Distance**: In the context of three-class (home win, draw, away win) prediction, a scoring rule is considered sensitive to distance if it takes into account the ordinal nature of the outcomes of the match. For example, in soccer, when a team is leading by one goal, it only takes one goal from the opponent to turn the match into a draw, and two goals for the opponent to secure a win. Consequently, the outcome space is ordered based on the goal differences.

Of the above-mentioned scoring rules, the Brier Score, RPS, and IGN/Log Loss are strictly proper, only IGN/Log Loss is local, and only RPS is sensitive to distance.

For the outcome prediction problem, it is essential to use a strictly proper scoring rule to incentivize the model to truthfully report its predictions [48, 16]. The objective is to develop a model that accurately reflects the true distribution of match outcomes.

Regarding the scoring rule for three-class prediction in soccer, Constantinou & Fenton [29] proposed using the RPS because of its sensitivity to distance, i.e., it takes into account the ordering in the outcomes (e.g., a home win is "closer" to a draw than it is to an away win). However, Wheatcroft [119] challenged this perspective, arguing that the sensitivity to distance does not add anything in terms of the usual aims of using scoring rules and that football outcomes may be drawn from an unknown distribution. Therefore, instead of evaluating predictions based on exact outcomes, it was suggested that it is more appropriate to consider the sample distribution of outcomes. In this context, the Ignorance (log loss) Score is a better choice due to its local property, which is more efficient compared to other non-local scoring rules.

While the sensitivity to distance property is still a topic of debate, the scoring rule that is selected should align with the objective of the prediction task.

### 1.5.4 Temporal Splitting Methods

The temporal order of matches should be preserved when evaluating results so that upcoming games are predicted based on past games only. One challenge in ML for sports match results prediction is that — unlike many applications of supervised machine learning techniques — the instances representing sports matches are temporally ordered. Furthermore, there is a hierarchy in terms of the way matches are structured, i.e., there are multiple matches in a particular round and multiple rounds that comprise a particular season. As a result, conventional cross-validation, which shuffles instances randomly, cannot be used since future matches may end up being incorrectly used to predict past matches [20, 18, 113]. Performance that appears extremely strong relative to other studies for an equivalent two- or three-class problem (e.g., a study reporting over 80% or 90% accuracy for a three-class problem when most other studies report between 45% to 55%) may be evidence that this mistake has been made in a study.[14] There is an option to preserve the order of the instances in the machine learning software WEKA (Figure 1.1), and code in programming languages such as Python and R can be written to ensure that the temporal order of matches is preserved when training and testing models.

---

[14] For example, [102] used 10-fold cross-validation in WEKA and reported obtaining accuracy of 99.56% with a decision tree model.

Deciding how to split match data into training and validation sets depends on the amount of data that the researcher has on hand, e.g., whether there is one season of match data or multiple seasons. Seasons further in the past become less relevant for predicting matches in the current or future seasons because of, e.g., changes in team rosters and team strengths. However, if player-level features are included, player changes from season to season can be accounted for. The time series nature of soccer match result data should thus be accounted for when researchers evaluate their models, e.g., using cross-validation for time series. Furthermore, any hierarchies that are present in terms of a particular league being made up of seasons that are comprised of rounds, which, in turn, consist of matches, should be considered. If the competition is structured such that teams sometimes play more than one match per round, this should also be taken into account, as should if teams are promoted and relegated at the end of seasons.

> **Important**

Any match result prediction project must ensure that only matches in the past are being used to predict current or future matches. Therefore, traditional cross-validation, which shuffles data instances randomly, should not be used.

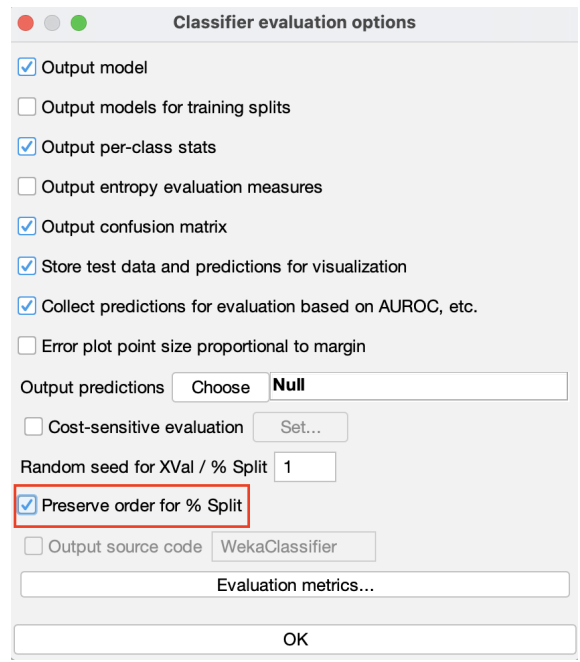

**Fig. 1.1** The "Preserve order for % Split" option can be set in WEKA to ensure that the temporal order of instances in a dataset is maintained so that future matches are not incorrectly used to predict historical matches.

## 1.6 Conclusions and Future Directions

Soccer match result prediction is inherently difficult due to soccer being a low-scoring sport with three outcomes and generally highly competitive leagues at the professional level. There is also an element of luck in any sport, including soccer [2], which is, of course, a major part of what attracts fans.

This chapter has aimed to provide a general overview of ML for soccer match result prediction and to act as a resource for researchers conducting future work in this domain. Available datasets, the types of models and features, and approaches to evaluating ML model performance were covered in this chapter, and while the chapter has not aimed to provide an exhaustive review of all studies in the area, it has given a contemporary overview by focusing, to a greater extent, on studies that have been published in recent years. The chapter has also not considered betting strategies to any large degree, which is a significant potential use case for ML match result prediction models in soccer, and researchers are encouraged to also familiarize themselves with the literature in this area if their models will be deployed for this purpose.

As discussed in subsection 1.2.2, the results from studies that have used the Open International Soccer Database [39] suggest that gradient-boosted tree models, e.g., CatBoost [93] and XGBoost[26], which are applied to feature sets consisting of soccer-specific ratings (pi-ratings [30] and Berrar ratings [8]), provide state-of-the-art performance for soccer match result prediction in the absence of match features apart from goals scored [8, 59, 96]. It remains unclear, however, whether ensemble methods are the best-performing models on datasets that contain match statistics in addition to goals (e.g., statistics contained in the European Soccer Database, football-data.co.uk) and/or external features, and also whether deep learning models can outperform ensemble models. On other datasets containing match statistics and other types of features, Random Forests have been competitive with [5] and even surpassed the performance gradient-boosted tree models [107, 1, 43]. Thus, comparing the performance of bagging methods (e.g., Random Forest) with boosting methods (e.g., CatBoost) and deep learning models on different datasets with varying features is an interesting avenue for further work. Furthermore, using the recently proposed Generalized Attack Performance (GAP) [118] ratings as model features is yet to be investigated. Rating systems themselves can also be used as predictive models by simply predicting the team with the higher rating as the winner; however, using the ratings as model features has been shown to provide better performance. Despite many sophisticated models having been developed, including hybrid approaches that combine statistical and machine learning models and rating systems [40, 74, 53], predictions derived from bookmaker odds still provide strong baseline performance [100, 5]. There are different ways in which expert knowledge can be incorporated (see the Appendix) into the model itself: through feature selection, or by using the match predictions of experts as a baseline (as mentioned above, former players appear to be better at predicting results [21]). Player or team ratings can be obtained from video games, constructed from performance indicator metrics, or by aggregating player ratings — computed using plus-minus ratings or VAEP — into a team-level rating.

The interpretability of models is of greater importance to certain groups such as sports coaches and performance analysts, and techniques including Random Forest feature importance, SHAP, and models new to the domain such as Alternating Decision Trees (see the Appendix) may be of use in identifying and interpreting performance indicator model features that are most relevant for winning. In the context of deep learning models, which are generally black-box, mimic learning [110] may be worth exploring to aid in their interpretability for match result prediction.

Predicting a different target variable, e.g., expected goals/xG, shot attempts, shots on target, etc., to predict the match result may be superior for prediction than using a three-class win/draw/loss outcome target variable. The best evaluation metric for probabilistic forecasting in this domain remains a subject of debate [30, 119]. Researchers should ensure that their model evaluation adequately accounts for the time series nature of soccer match result data and its hierarchical nature (i.e., match, round, season).

While all existing rating systems account for historical results — as well as accounting for the venue (home/away) and creating separate offensive and defensive ratings in the case of the soccer-specific rating systems — developing new rating systems that incorporate additional factors is an avenue for further research. For instance, rating systems could incorporate information derived from, e.g., event log data, spatiotemporal tracking data, player ratings, player/team attributes and performance indicators, passing networks, inter-player chemistry, and even information from social media. Streak and form features could potentially be engineered at both the player (hot shoe) and team levels (form and streaks in terms of match outcomes [5]) and their value as model features compared.

**Acknowledgements** This chapter was partly supported by JSPS KAKENHI (grant number 20H04075) and JST Presto (grant number JPMJPR20CA).

# Appendix

## Incorporating Expert Knowledge

The incorporation of domain expert knowledge has been a key area of enquiry for some time. Expert knowledge can be incorporated into the modelling process itself: e.g., Joseph, Fenton & Neil [68] did so in constructing their Bayesian Network. Alternatively, expert knowledge can be incorporated by obtaining expert predictions as a benchmark with which to compare model prediction performance (see also subsection 1.5.2). Another way in which expert domain knowledge of the sport can be incorporated is by asking experts to select features. Hucaljuk and Rakipović [62] considered 96 matches from the European Champions League and compared the performance of ML models that included features selected by feature selection algorithms with models that included features selected by experts. The feature se-lection algorithms produced a basic feature set of 20 features in their initial feature set, which they compared with a feature set consisting of this initial set plus the expert-selected features. With their particular models and on their specific dataset, however, the expert-selected features were not found to yield any improvement in model performance. More recently, Beal et al. [6] developed a benchmark dataset and results from Natural Language Processing and ML models for soccer match result prediction, and the performance of these models was compared with statistical models. The authors provided a prediction accuracy baseline that utilizes match statistics and match preview articles written by sports journalists in The Guardian (London, United Kingdom) over six English Premier League seasons.

### Elo K-factor Selection Techniques Described Online

The World Football Elo Ratings Website (eloratings.net) [103] provides Elo ratings of national teams, and uses a higher value of $K$ for matches that are of greater importance. For example, $K = 60$ for World Cup finals matches and $K = 20$ for International friendlies. The $K$-factor is also adjusted based on the goal margin. Specifically, the $K$-factor is increased by 1/2 for a win by 2 goals, by 3/4 for a win by 3 goals, and by $\frac{3}{4} + \frac{(M-3)}{8}$ if the team wins by 4 or more goals, where $M$ denotes the goal margin.

The Football Database website (footballdatabase.com) provides Elo ratings for club teams from various leagues around the world, also applying different values of $K$ based on the league or importance of the match.[15] The $K$-factor is multiplied by a value, $G$, which accounts for the goal margin. In particular, $G = 1$ if the match was drawn, $G = 1.5$ if the goal margin was 1 or 2 goals, and $G = \frac{11+M}{8}$ if the goal margin was 3 or more, where $M$ again denotes the goals margin.

### 2023 Soccer Prediction Challenge

A subsequent Soccer Prediction Challenge, using a similar dataset to the 2017 challenge, was held in 2023.[16] However, unlike the 2017 competition, the 2023 version required two tasks. First, predicting match results in terms of the exact goals scored by each team, and second, as per the 2017 challenge, predicting match results by computing the probabilities for a win, draw, and loss. The exact score prediction models were evaluated using the Root-Mean-Squared-Error metric. Similar to the 2017 challenge, studies will be submitted to the Machine Learning (Springer) journal; however, these studies were not available at the time the current chapter was written.

---

[15] See footballdatabase.com/methodology.php for further details.

[16] https://sites.google.com/view/2023soccerpredictionchallenge

### Alternating Decision Trees (ADTrees)

An interesting boosted-tree model that is relatively unexplored in sports result prediction is the Alternating Decision Tree (ADTree) model [46]. As opposed to XGBoost and CatBoost, both of which use gradient boosting and gradient descent optimization, ADTree uses the well-known AdaBoost [47] algorithm. AdaBoost is used to grow the ADTree, with each iteration of AdaBoost adding a branch to the ADTree. The general idea of AdaBoost is to iteratively place more weight on instances that were previously incorrectly classified in a previous iteration. The ADTree algorithm combines the accuracy-increasing benefits of boosting while producing an interpretable decision tree structure as the final ADTree model. Given the interpretability of the ADTree, an enhanced version of ADTree that can compete with the likes of CatBoost would be a great asset to the sports match results prediction domain.[17]

---

[17] Bunker, Yeung, Susnjak, Espie & Fujii [19] recently found that ADTrees performed well in predicting professional ATP tennis match results when applied to the difference in average betting odds across 11 different bookmaker companies (the companies included in the betting odds data on tennis-data.co.uk). Betting odds, as mentioned before, can be considered a rating [122], so this is another example of boosted-tree models, applied to rating features, being effective for sports match result prediction.